\documentclass[sigconf]{acmart}

\usepackage[true]{anonymous-acm}
\usepackage{titlesec}
\usepackage[utf8]{inputenc}
\usepackage{amsmath}
\usepackage[ruled,vlined]{algorithm2e}
\usepackage[capitalise]{cleveref}
\usepackage{booktabs}
\usepackage{pifont}
\usepackage{adjustbox}
\usepackage{subcaption}
\usepackage{graphicx}
\usepackage{cleveref}
\usepackage{threeparttable}

\usepackage{balance} 
\usepackage{relsize} 
\usepackage{amsmath}

\AtBeginDocument{%
  \providecommand\BibTeX{{%
    \normalfont B\kern-0.5em{\scshape i\kern-0.25em b}\kern-0.8em\TeX}}}
 
 \acmYear{2021}\copyrightyear{2021}
 \acmConference[]{ }

\begin{document}


\title[Real-world challenges]{Real-world challenges for multi-agent reinforcement learning in grid-interactive buildings}

\author{Kingsley Nweye and Zoltan Nagy}

\email{{nweye, nagy}@utexas.edu}
\affiliation{
  \institution{Department of Civil, Architectural and Env Engineering\\
  The University of Texas at Austin}
  \streetaddress{301 E. Dean Keeton St.}
  \city{Austin}
  \state{Texas}
  \country{USA}
  \postcode{78712-1700}
}

\author{Bo Liu and Peter Stone}
\email{{bliu,pstone}@cs.utexas.edu}
\affiliation{
  \institution{Department of Computer Science\\
  The University of Texas at Austin}
  \streetaddress{2317 Speedway}
  \city{Austin}
  \state{Texas}
  \country{USA}
  \postcode{78712-1700}
}

\renewcommand{\shortauthors}{Nweye et al}

\begin{abstract}
Building upon prior research that highlighted the need for standardizing environments for building control research, and inspired by recently introduced challenges for real life reinforcement learning control, here we propose a non-exhaustive set of nine real world challenges for reinforcement learning control in grid-interactive buildings. We argue that research in this area should be expressed in this framework in addition to providing a standardized environment for repeatability. Advanced controllers such as model predictive control and reinforcement learning (RL) control have both advantages and disadvantages that prevent them from being implemented in real world problems. Comparisons between the two are rare, and often biased. By focusing on the challenges, we can investigate the performance of the controllers under a variety of situations and generate a fair comparison. As a demonstration, we implement the offline learning challenge in CityLearn and study the impact of different levels of domain knowledge and complexity of RL algorithms. We show that the sequence of operations utilized in a rule based controller (RBC) used for offline training affects the performance of the RL agents when evaluated on a set of four energy flexibility metrics. Longer offline learning from an optimized RBC leads to improved performance in the long run. RL agents that learn from a simplified RBC risk poorer performance as the offline learning period increases. We also observe no impact on performance from information sharing amongst agents. We call for a more interdisciplinary effort of the research community to address the real world challenges, and unlock the potential of grid-interactive building controllers.
\end{abstract}


\maketitle
\section{Introduction} \label{sec:introduction}
\begin{figure}[t]
     \centering
     \includegraphics[width=\columnwidth]{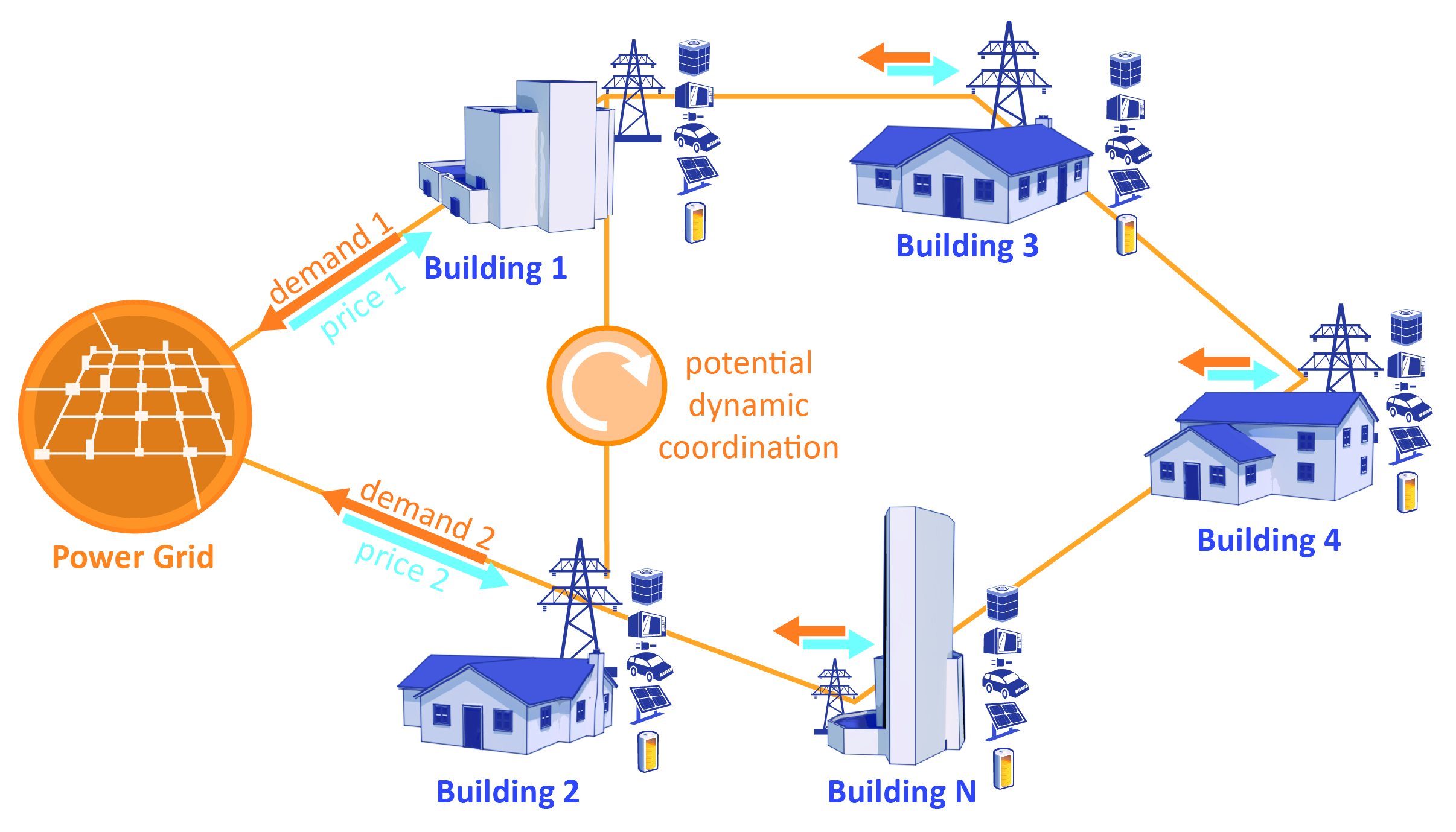}
     \caption{Grid-interactive buildings}
     \label{fig:demand_response}
 \end{figure}

Buildings account for $\approx\!40\%$ of the global energy consumption and $\approx\!30\%$ of the associated greenhouse gas emissions, while also offering a 50---90\% CO$_2$ mitigation potential~\cite{ipcc14}. Optimal decarbonization requires electrification of end-uses and concomitant decarbonization of electricity supply, efficient use of electricity for lighting, space heating, cooling and ventilation (HVAC), and domestic hot water generation, and upgrade of the thermal properties of buildings~\cite{Lanham2018}. A major driver for grid decarbonization is integration of renewable energy systems (RES) into the grid (supply) and, photovoltaics (PV) and solar-thermal collectors into residential and commercial buildings (demand). Electric vehicles (EVs), with their storage capacity and inherent connectivity, hold a great potential for integration with buildings~\cite{Mohagheghi2010DemandSystem}. However, this grid-building integration must be carefully managed during operation to ensure reliability and stability of the grid~\cite{Vazquez-Canteli2018d, Dupont2014ImpactStudy,DOE2021} (Fig.\ref{fig:demand_response}). 

Demand response (DR) as an energy-management strategy enables end-consumers to provide the grid with more flexibility by reducing their energy consumption through load curtailment, shifting their energy consumption over time, or generating and storing energy at certain times (\cref{fig:demand_response}). In exchange, consumers typically receive a reduction of their energy bill~\cite{Siano2014}. HVAC can contribute to load curtailment events by modifying the temperature set points, participating in load shifting by pre-heating or pre-cooling the buildings~\cite{Bruninx2013Short-termSimulations} (passive energy storage), or by directly storing thermal energy in an energy storage system (active energy storage). Thermostats with DR functionality can provide energy savings to residential customers by allowing electricity retailing companies to adjust set-points during peak-demand events. 
Widespread integration of communication technologies allows all involved systems (PV, HVAC, storage, EVs, thermostats, etc.) to exchange information on their operation, leading to the concept of \emph{smart cities}, allowing cities to achieve energy savings, and become more sustainable~\cite{Chourabi2011UnderstandingFramework}.

Advanced control systems can be a major driver for DR by automating the operation of energy systems, while adapting to individual characteristics of occupants and buildings. However, for DR to be effective, loads must be controlled in a responsive, adaptive and intelligent way. When all the electrical loads react simultaneously to the same price signals, aggregated electricity peaks could be shifted rather than shaved. Therefore, there is a need for more efficient and effective ways of coordinating the response of all the technologies described above.

Advanced control algorithms such as model predictive control (MPC)~\cite{Drgona2020} and deep reinforcement learning (RL)~\cite{Wang2020} have been proposed for a variety of building control applications. While both  methods have their disadvantages, e.g., MPC requiring a model while RL being data intensive, spectacular applications and results have been presented in the past several years. In addition, recently, hybrid methods, based on physics constrained neural networks for models have begun to emerge~\cite{Drgona2021}. 


In contrast to MPC, RL is an adaptive and potentially model-free control algorithm that can take advantage of both real-time and historical data to provide DR capabilities. RL is an agent-based machine learning algorithm in which the agent learns optimal actions via interaction with its environment~\cite{RichardS.SuttonandAndrewG.Barto2018a,Nagy2018}. In contrast to supervised learning, the agent does not receive large amounts of labelled data to learn from. In contrast to unsupervised learning, the agent receives delayed feedback from the environment. In brief, for a given input, the agent chooses to perform a certain action. It then observes an immediate or delayed reward signal from the environment, and uses it to modify its knowledge on which action is best to choose under given circumstances. 


A major challenge for RL in demand response is the ability to compare algorithm performance~\cite{Vazquez-Canteli2018d}. As argued in~\cite{Wolfle2020}, a 
\begin{quote}
    \emph{shared collection of representative environments} [needs to be established in order to] \emph{systematically compare and contrast [...] building optimization algorithms}. 
\end{quote}
Building on~\cite{Wolfle2020}, and inspired by~\cite{Dulac-Arnold2021}, the purpose of this paper is twofold. First, we introduce and discuss specific real world challenges for grid-interactive buildings that our community should be focusing on. Second, we demonstrate one particular challenge using the CityLearn gym environment~\cite{Vazquez-Canteli2019}. 

This paper is organized as follows. \cref{sec:real_world_challenges} presents nine real world challenges for grid-interactive building, while \cref{sec:background} provides background on reinforcement learning and CityLearn. In \cref{sec:methodology}, we provide a framework towards addressing one of the introduced challenges and present our results from addressing said challenge using a case study data set. A discussion of the results and conclusion follow in \cref{sec:discussion} and \cref{sec:conclusion}.

\section{Real-world challenges} \label{sec:real_world_challenges}
Dulac-Arnold et al. provide nine real-world challenges for RL~\cite{Dulac-Arnold2021}. Their challenges are not suitable to evaluate grid-interactive buildings, as they are based on small scale environments without the necessary domain knowledge or context. In the following, we present a first set of challenges; we provide the description of~\cite{Dulac-Arnold2021} in \emph{italics}.


\begin{itemize}
    \item[\textbf{C1:}] \emph{Being able to learn on live systems from limited samples:} In this challenge, the controller is initialized randomly and has to learn to perform only based on the samples it observes. The sample size can be artificially reduced by presenting the controller only with a subset of the data, e.g., every 3 hours instead of every 15 min. The algorithms can be evaluated on how quickly in terms of time or sample number they converge, and how stable their exploration is. Conversely, we can evaluate the trade-off between data requirement and controller performance. 
    
    \item[\textbf{C2:}] \emph{Dealing with unknown and potentially large delays in the system actuators, sensors, or feedback} The thermal dynamics of buildings are such that the effects of controller actions to adjust the HVAC systems are observed in delays. This has implications for, e.g., pre-cooling/heating of buildings to take advantage of the thermal mass of buildings. The controller need to implicitly and automatically learn the dynamics of the building. Challenge datasets with different thermal mass from light to heavy should be created, and the converged controller should be compared to understand the relationship between longer delays in feedback (higher thermal mass) and controller performance.
    
    \item[\textbf{C3:}] \emph{Learning and acting in high-dimensional state and action spaces.} 
    This challenge addresses the scalability of a proposed controller. As buildings can inherently have a large state-action space, controller can be evaluated on specific subsets of them to understand how the performance changes. In the case of controlling multiple buildings (or multiple zones within a building), scalability refers to essentially increasing the number of buildings (or zones) and observe the control performance.
    
    \item[\textbf{C4:}] \emph{Reasoning about system constraints that should never or rarely be violated} This is a central challenge, as building control problems are indeed often presented as balancing between reducing energy use while maintaining comfortable conditions. Other constraints in the energy system are operational, such as ensuring a minimum state of charge, maintaining operational temperatures within limits, etc. The algorithms should be  evaluated on both the number of violations during the learning process and for the converged policy. Integration of constraint violation into the objective function is addressed in C6 below.
    
    \item[\textbf{C5:}] \emph{Interacting with systems that are partially observable, which can alternatively be viewed as systems that are non-stationary or stochastic.} This challenge has two parts. In the first part, observations can be modified to 
    contain failures (sensor noise, missing data, etc.), which can be common in any real life systems, like buildings and HVAC systems. We then observe the performance of the algorithms for various levels of the failures (more noise, more missing data). In the second part, we can observe how a controller performs on a perturbed system. Perturbations can consist of retrofit measures on buildings (improving envelop or windows), improving equipment, changed occupant behavior or different climate. We can then judge the algorithms on their ability to perform their previously learned policy on the perturbed system. 
    
    \item[\textbf{C6:}] \emph{Learning from multiple or poorly specified objective functions.} Energy management in buildings in inherently multi-objective, especially when considering multiple zones or multiple buildings. Another example is when there is a global objective (overall building energy use) as well as multiple local objectives (equipment operation). As mentioned in C4, constraints can be incorporated into the objective function directly. When evaluating the controller performance, the individual objectives should be separated to allow for a fair comparison. 
    
    \item[\textbf{C7:}] \emph{Being able to provide actions quickly, especially for systems requiring low latencies.} Latency is a delay in executing a control action after acquiring a measurement due to long computational time. Latencies in real life systems can occur if the system dynamics are fast or the computational times long. A practical example for smart buildings and microgrids is if the computation is taking place in the cloud, adding also data transfer to the execution time, which can be exacerbated by connectivity issues. To observe the impact of latency, time-step delays of various lengths should be included into the control execution and the impact on their performance should be evaluated.
    
    \item[\textbf{C8:}] \emph{Training off-line from the fixed logs of an external behavior policy.} The challenge here is to learn a control law from data generated by a suboptimal reference controller, e.g., a rule based controller, which is often available, essentially a system log. In addition to the control environment, datasets of various size, e.g., two weeks, one month, six months should be provided that are generated with a known reference rule based controller. Then, the controllers can be evaluated on the ability to improve these baselines.
    
    \item[\textbf{C9:}] \emph{Providing system operators with explainable policies.} Here we deviate from the description in~\cite{Dulac-Arnold2021} who propose to generate figures to improve the interpretability of the results. Rather, for the building context, what is needed is that the control actions can be explained simply to building managers. Advances in explainable AI are needed, and algorithms that might perform suboptimally, yet are easier to explain are favored as they are more likely to get accepted, and thus implemented. A consensus between modelers and system operators on the standards and outcomes of a control law could be established to facilitate effective communication amongst invested parties.
\end{itemize}

Each of the aforementioned challenges require unique experimental designs within a simulation environment to adequately study and quantify the factors that affect their resolution. We demonstrate challenge \textbf{C8} using the CityLearn environment \cite{Vazquez-Canteli2019} in \cref{sec:methodology}.

\section{Background} \label{sec:background}
We provide a background on reinforcement learning and multi-agent reinforcement learning. Detailed introductions can be found in standard textbooks~\cite{RichardS.SuttonandAndrewG.Barto2018a}.

\subsection{Reinforcement Learning}
In reinforcement learning (RL), an agent interacts with an environment to maximize the reward it receives. RL is usually formulated as a Markov decision process (MDP). An MDP $\mathcal{M}$ is a tuple $\mathcal{M} = (\mathcal{S}, \mathcal{A}, \mathcal{T}, \gamma, R)$. $\mathcal{S}$ and $\mathcal{A}$ are the state and action spaces for the agent. At time step $t$, the agent is located at a state $s_t \in \mathcal{S}$. After taking an action $a_t \in \mathcal{A}$, the agent will be transitioned to the next state $s_{t+1} \sim \mathcal{T}(\cdot \mid s_t, a_t)$, where $\mathcal{T}$ denotes the transition probability and is usually hidden from the agents. Moreover, the agent receives a scalar reward $r_t \sim R(s_t, a_t)$. The overall objective of RL is to find a policy $\pi: \mathcal{S} \rightarrow \mathcal{A}$ that maximizes the expected cumulative return:
\begin{equation}
    \label{eq:rl-obj}
    \max_\pi \mathbb{E}_{s_t, a_t \sim \pi(\cdot \mid s_t)}\bigg[\sum_{t=0}^\infty \gamma^t r_t\bigg]. 
\end{equation}

It has been shown that given any stationary policy $\pi$, the above objective will converge to a value based on which state the agent starts from. Specifically, we have the value of a policy defined as:
\begin{equation}
    V^\pi(s) = \mathbb{E}_{s_0 = s, \pi}\bigg[\sum_{t=0}^\infty \gamma^t r(s_t, a_t)\bigg],
\end{equation}
where $r(s_t, a_t) = r_t \sim R(s_t, a_t)$ and we use $\mathbb{E}_{\pi}$ to denote that the expectation is taken over the trajectories sampled from the policy $\pi$. Similarly, we can define the action-value function:
\begin{equation}
    Q^\pi(s, a) = \mathbb{E}_{s_0=s, a_0=a, \pi}\bigg[\sum_{t=0}^\infty \gamma^t r(s_t, a_t)\bigg].
\end{equation}

The RL objective in Eq.~\ref{eq:rl-obj} is therefore equivalent to
\begin{equation}
    \max_\pi V^\pi(s), \forall s.
\end{equation}

To optimize the above objective, there are typically two types of RL algorithms: value-based and policy-based. The value-based algorithms are based on the well-known Bellman equation of the action-value function. Denote the optimal action-value function as $Q^*$, then it is known that for $Q^*$, it satisfies
\begin{equation}
    Q^*(s, a) = r(s, a) + \gamma \mathbb{E}_{s' \sim \mathcal{T}(\cdot \mid s,a)} \max_{a'} Q^*(s', a').
\end{equation}
By minimizing the difference between the left and right-hand sides of the above equation, we reach the Q-learning algorithm\cite{Watkins1992}. 



 \subsection{Multiagent Reinforcement Learning}
\begin{figure}[t]
     \centering
     \includegraphics[width=\columnwidth]{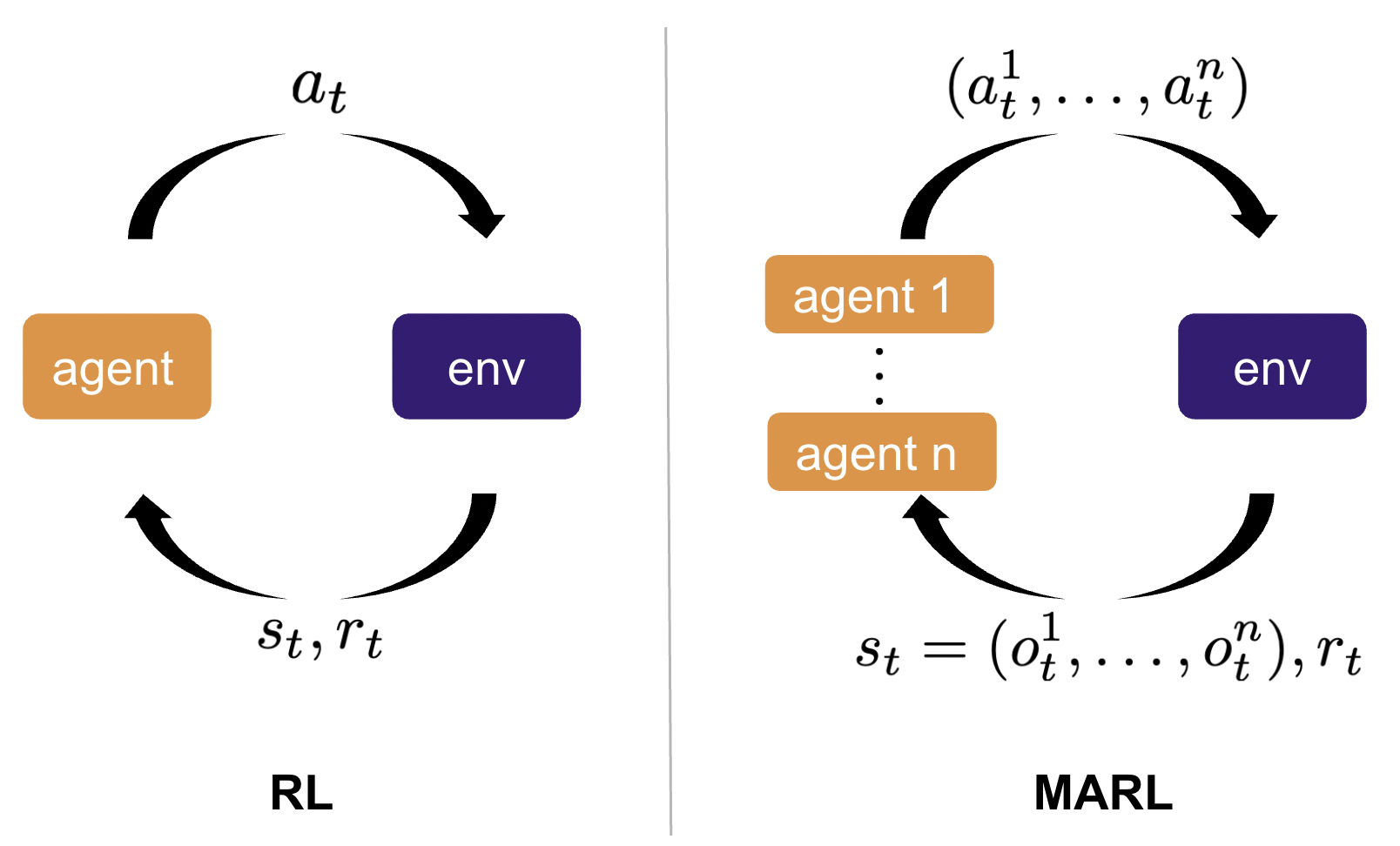}
     \caption{The pipeline of RL and MARL.}
     \label{fig:marl}
 \end{figure}

 Multiagent reinforcement learning (MARL) extends RL to the setup involving multiple agents. The general MARL framework includes the cooperative setup, the competitive setup and the mixture of the two. In this work, we focus on the cooperative setup because the main objective is to coordinate buildings to flatten the electricity demand curve, which is a shared objective for all agents. To summarize, the MARL problem we consider in this work is also formulated as a Markov decision process represented by the tuple $\mathcal{M} = (\mathcal{S}, \mathcal{A}, \mathcal{T}, \gamma, R)$. The major differences are: 1) the action space now includes the joint actions of all agents, i.e. $\mathcal{A} = \mathcal{U}^1 \times \mathcal{U}^2 \dots \times \mathcal{U}^n$, where $\mathcal{U}^i$ is the action space of the $i^{\textrm{th}}$ agent. 2) the state space $\mathcal{S} = \mathcal{O}^1 \times \mathcal{O}^2 \dots \times \mathcal{O}^n$, where $\mathcal{O}^i$ is the observation of the $i^{\textrm{th}}$ agent. The pipeline of RL and MARL are summarized in \cref{fig:marl}.
 
 Although in principle a multiagent problem can be regarded as a single agent problem where a centralized agent chooses actions for all agents, it is both computationally expensive and hard to train in practice, as the state and action space grow dramatically with the number of agents. Therefore, in practice, decentralized algorithms that learn a decision module for each agent is a more practical approach. On the other hand, a fully decentralized algorithm where agents are not aware of other agents' policies might result in poor coordination. 
 
 \subsection{CityLearn}
\begin{figure}
     \centering
     \includegraphics[width=\columnwidth]{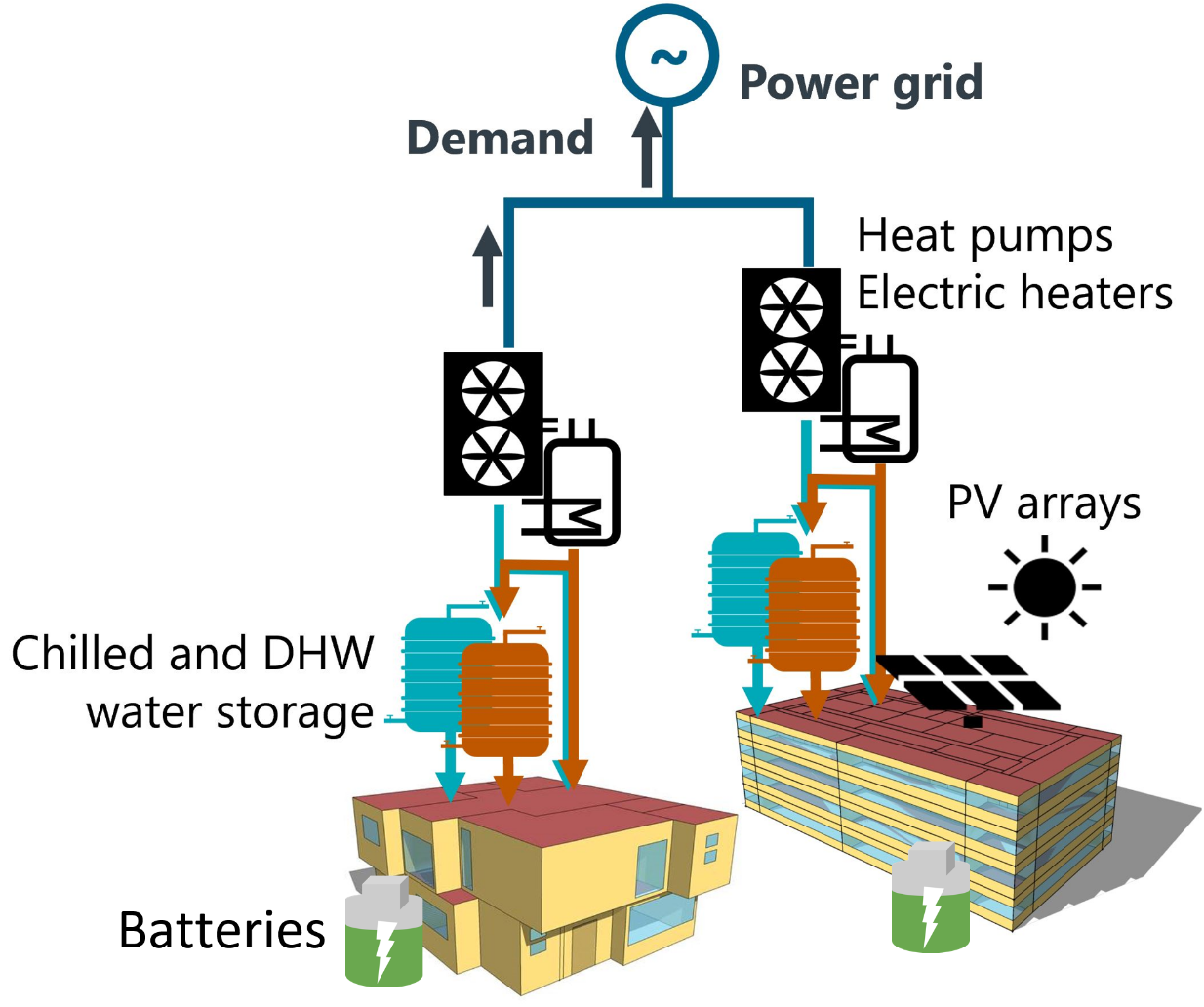}
     \caption{CityLearn overview.}
     \label{fig:citylearn}
 \end{figure}

CityLearn is an OpenAI Gym environment for the easy implementation of RL agents in a demand response setting to reshape the aggregated curve of electricity demand by controlling the energy storage of a diverse set of buildings in a district~\cite{Vazquez-Canteli2019,citylearnArxiv, citylearngithub}. Its main objective is to facilitate and standardize the evaluation of RL agents, such that it enables benchmarking of different algorithms. CityLearn includes energy models of air-to-water heat pumps, electric heaters, chilled water, domestic hot water (DHW) and electricity energy storage devices as shown in \cref{fig:citylearn}. In each building, the air-to-water heat pump is used to meet the cooling demand and an electric heater is used to meet DHW heating demand. Buildings could also possess a combination of chilled water, DHW and electricity storage devices to offset cooling, DHW heating and electricity demand from the grid. Chilled water and DHW storage capacities are represented as a multiple of the hours the storage device can satisfy the maximum annual hourly cooling or DHW demand if fully charged. All these devices, together with other electric equipment and appliances (non-shiftable loads) consume electricity from the main grid. Photovoltaic (PV) system may be included in the buildings' energy systems to offset part of this electricity consumption by allowing the buildings to generate their own electricity.

The RL agents control the storage of chilled water, DHW and electricity by deciding how much cooling, heating and electrical energy to store or release at any given time. CityLearn guarantees that, at any time, the heating and cooling energy demand of the building are satisfied regardless of the actions of the controller by utilizing pre-computed energy loads of the buildings, which include space cooling, dehumidification, appliances, domestic hot water (DHW), and solar generation. The backup controller guarantees that the energy supply devices prioritize satisfying the energy demand of the building before storing any additional energy. 

CityLearn has been used extensively as a reference environment to demonstrate incentive-based demand response\cite{Deltetto2021}, collaborative demand response\cite{Glatt2021}, coordinated energy management\cite{Pinto2021, Kathirgamanathan2020}, or benchmarking RL algorithms\cite{Dhamankar2020, Qin2021}. Here we use the dataset that was made public for the CityLearn Challenge 2021~\cite{Nagy2021}. It consists of nine DOE prototype buildings: one medium office (ID=1), one fast-food restaurant (ID=2), one standalone retail (ID=3), one strip mall retail (ID=4), and five medium multifamily buildings (ID=5--9). The energy demand for each building has been pre-simulated in EnergyPlus using 2014--2017 actual meteorological year weather data\textanon{ for Austin, TX}. Their cooling, DHW and electricity  storage capacities, as well as PV capacities, are shown in \cref{tab:building_metadata}.

\begin{table}
\centering
\caption{Chilled water, domestic hot water (DHW) and electricity  storage and, photovoltaic (PV) capacities per building. The unit of measurement for chilled water and DHW storage capacity is the hours of maximum annual hourly cooling and DHW demand that can be satisfied on full charge.}
\label{tab:building_metadata}
\begin{tabular}{lrrrr}
\toprule
{} & Chilled water & DHW & Electricity & PV \\
ID & Storage (h) & Storage (h) & Storage (kWh) & (kW)  \\
\midrule
1  & 2 & 2 & 140 & 120 \\
2  & 3 & 3 & 80 & 0 \\
3  & 2 & 0 & 50 & 0 \\
4  & 1.5 & 0 & 75 & 40 \\
5  & 3.5 & 1.5 & 50 & 25 \\
6  & 1.5 & 3 & 30 & 20 \\
7  & 2 & 2 & 40 & 0 \\
8  & 3 & 3 & 30 & 0 \\
9  & 3 & 3 & 35 & 0 \\
\bottomrule
\end{tabular}
\end{table}

\section{Offline Learning Challenge (C8)} \label{sec:methodology}
Here, we provide a framework for studying \textbf{C8}. Specifically, we compare two RL control approaches, (1) independent, uncoordinated SAC agents (see \cref{subsubsec:sac}), and (2) the MARLISA algorithm for coordinating the agents (see \cref{subsubsec:marlisa}).
We investigate the agents' behavior with respect to varied periods of offline training from a rule based controller (RBC). Our central hypothesis is that \textit{a longer offline training period results in better performance, since the agents will have more existing knowledge of what ideal actions could resemble by the time they come online}. We study the effect of RBC controller's sequence of operation on our hypothesis by evaluating the controllers on a set of performance metrics or cost functions.

\subsection{Agent \& Reward Design}
\subsubsection{Independent SAC agents} \label{subsubsec:sac}
To control environments that have continuous states and actions, tabular Q-learning is not practical, as it suffers the curse of dimensionality. Actor-critic RL methods use artificial neural networks to generalize across the state-action space. The actor network maps the current states to the actions that it estimates to be optimal. Then, the critic network evaluates those actions by mapping them, together with the states under which they were taken, to the Q-values.

Soft actor-critic (SAC) is a model-free off-policy RL algorithm \cite{Haarnoja2018}. As an off-policy method, SAC can reuse experience and learn from fewer samples. SAC is based on three key elements: an actor-critic architecture, off-policy updates, and entropy maximization for efficient exploration and stable training. SAC learns three different functions: the actor (policy), the critic (soft Q-function), and the value function $V$. For more details about SAC, we refer the reader to \cite{Haarnoja2018SoftAA}.

\begin{equation}
    r_{i}^{\textrm{SAC}}(t) = \textrm{min}\left(0, e_{i}(t)\right)
    \label{eqn:sac_reward}
\end{equation}

We use the reward $r_{i}^{\textrm{SAC}}(t)$ (\cref{eqn:sac_reward}) for the independent SAC RL agents. It is a single-agent reward whose value only depends on the net
electricity consumption $e_{i}(t)$ of the agent $i$ at timestep $t$. $e_{i}(t) < 0$ if the
building is consuming more electricity than it generates, and $e_{i}(t) > 0$
if the building is self-sufficient at that time and generates excess
electricity.

\subsubsection{MARLISA RL Agents}
\label{subsubsec:marlisa}
MARLISA is built on the SAC algorithm and allows for coordination of the agents through reward sharing, collective rewards, as well as mutual sharing of some information~\cite{Vazquez-Canteli2020}. The agents predict their own future electricity consumption and share this information with each other, following a leader-follower schema. In an iterative process, each agent converges to selecting an action before the action is implemented.

\begin{equation}
    r_{i}^{\textrm{MARL}}(t) = -\textrm{sign}(e_{i}(t)) \cdot 0.01 \cdot e_{i}(t)^{2} \cdot \textrm{min}\left(0, \sum_{i=0}^{n}{e_{i}(t)}\right)
    \label{eqn:marl_reward}
\end{equation}

$r_{i}^{\textrm{MARL}}(t)$ defined in \cref{eqn:marl_reward} is the MARLISA RL agents' reward function. It is a combination of the building level net electricity consumption $e_{i}(t)$ and the collective component $\sum{e_{i}(t)}$ , i.e., the total net electricity consumption of the entire district at timestep $t$, and is used to share information between the agents, which rewards them for reducing the coordinated energy demand.

\subsubsection{RBC}
We assumed no detailed knowledge of the energy profile of each building and developed two variations of RBC sequence of operation where, RBC\textsubscript{Basic} (\cref{alg:basic_rbc}) mimics a simplified logic and RBC\textsubscript{Optimized} (\cref{alg:optimized_rbc}) is informed by domain knowledge. For both sequences of operation, the input is the hour of the day, $h$ and timestep, $t$ and, the output is the charge/discharge action, $a{t}$ for chilled water, DHW or electricity storage. The RBC is tuned to act greedily in every building and use its storage capacity to reduce energy consumption by storing more energy during the night (when the coefficient of performance of the heat pumps is higher) and release it during the day. We also use the RBC to normalize the RL agents' performance metrics.

\begin{algorithm}
    \SetAlgoLined
    \KwIn{$h$,$t$}
    \KwOut{$a(t)$}
    \uIf{$9 \le h \le 21$}{
        a(t) = -0.08;
    }\Else{
        a(t) = 0.091;
    }
    \caption{RBC\textsubscript{Basic} sequence of operation.}
    \label{alg:basic_rbc}
\end{algorithm}

\begin{algorithm}
    \SetAlgoLined
    \KwIn{$h$,$t$}
    \KwOut{$a(t)$}
    \uIf{$1 \le h \le 6$}{ 
        a(t) = 0.05532;
    }\uElseIf{$7 \le h \le 15$}{
        a(t) = -0.02;
    }\uElseIf{$16 \le h \le 18$}{
        a(t) = -0.044;
    }\uElseIf{$19 \le h \le 22$}{
        a(t) = -0.024;
    }\Else{
        a(t) = 0.034;
    }
    \caption{RBC\textsubscript{Optimized} sequence of operation.}
    \label{alg:optimized_rbc}
\end{algorithm}

\subsection{Action-Space Design}
The action space per building is determined by the number of available energy storage systems to control, including the chilled water, DHW and electricity storage systems. Hence, the action space is bounded at $n * 3$ for a district of $n$ buildings that each possess the 3 storage systems. The action value is bounded between -1 and 1 where positive and negative values are charge and discharge control actions respectively.

\subsection{State-Space Design}
The available state space is made up of 27 observable temporal, weather, district, and building variables which are summarized in \cref{tab:state_space}. The storage system state of charge (SOC) states are conditionally available in each building. Meanwhile, the RBC controllers utilize only the \textit{hour} state in determining the control action.

\begin{table}[h]
    \centering
    \caption{The unified state space for all agents.}
    \label{tab:state_space}
    \begin{tabular}{llr}
        \toprule
        State & Unit \\
        \midrule
        \textbf{Temporal} & \\
        Month & - \\
        Day   & -  \\
        Hour  & - \\
        \textbf{Weather} & \\
        Outdoor dry-bulb temperature  & $^{\circ}$C  \\
        Outdoor dry-bulb temperature (6-hour forecast) & $^{\circ}$C \\
        Outdoor dry-bulb temperature (12-hour forecast) & $^{\circ}$C \\
        Outdoor dry-bulb temperature (24-hour forecast) & $^{\circ}$C \\
        Outdoor relative humidity & \% \\
        Outdoor relative humidity (6-hour forecast) & \% \\
        Outdoor relative humidity (12-hour forecast) & \% \\
        Outdoor relative humidity (24-hour forecast) & \% \\
        Diffuse solar irradiance & W/m\textsuperscript{2} \\
        Diffuse solar irradiance (6-hour forecast) & W/m\textsuperscript{2} \\
        Diffuse solar irradiance (12-hour forecast) & W/m\textsuperscript{2} \\
        Diffuse solar irradiance (24-hour forecast) & W/m\textsuperscript{2} \\
        Direct solar irradiance & W/m\textsuperscript{2} \\
        Direct solar irradiance (6-hour forecast) & W/m\textsuperscript{2} \\
        Direct solar irradiance (12-hour forecast) & W/m\textsuperscript{2} \\
        Direct solar irradiance (24-hour forecast) & W/m\textsuperscript{2} \\
        \textbf{District} & \\
        Net electricity consumption & kWh \\
        Carbon intensity & kg\textsubscript{CO\textsubscript{2}}/kWh\\
        \textbf{Building} & \\
        Indoor dry-bulb temperature & $^{\circ}$C \\
        Indoor relative humidity & \% \\
        Non-shiftable load & kWh \\
        Solar generation & W \\
        Chilled water storage state-of-charge & - \\
        Domestic hot water storage state-of-charge  & - \\
        Electricity storage state-of-charge & - \\
        \bottomrule
    \end{tabular}
\end{table}

\subsection{Performance Metrics/Cost Functions}
We evaluate the agents' performance on a set of cost functions that quantify the collective district's energy flexibility as follows:

\subsubsection{Average Daily Peak} is the average of all the daily peaks of the 365 days of the year and is calculated using the net energy demand of the whole district of buildings as defined by Equation \ref{eqn:cost_function-average_daily_peak} where $d$ is the day of the year and $i$ is the number of timesteps in a day. In our application, $i=24$ for an hourly resolution.
\begin{equation}
    \textrm{Average Daily Peak} = \left(\sum_{d=0}^{364}{\textrm{max}(Q_{i \times d},\dots,Q_{i \times (1 + d) - 1})}\right) \times\frac{1}{365}
    \label{eqn:cost_function-average_daily_peak}
\end{equation}

\subsubsection{Load Factor} is the difference between 1 and the ratio of average monthly demand to monthly peak demand defined by Equation  \ref{eqn:cost_function-1_load_factor} where $Q_t$ is the net electric consumption at timestep $t$ in the $m^{\textrm{th}}$ month and $k$ is the total number of timesteps per month. $k=730$ in our application where we use an hourly timestep resolution.
\begin{equation}
    1 - \textrm{Load Factor} = \left(\mathlarger{\mathlarger{\sum}}_{m=0}^{11}{
            1 - \frac
                    {\sum_{t=m \times k}^{k \times (1+m)-1}{Q_t}}
                    {k \times \textrm{max}(Q_t,\dots,Q_{k \times (1+m)-1})}
    }\right) \times\frac{1}{12}
    \label{eqn:cost_function-1_load_factor}
\end{equation}

\subsubsection{Net Electricity Demand} is defined by Equation \ref{eqn:cost_function-net_electricity_demand} as the sum of \textit{positive} net electricity demand because the objective is to minimize the energy consumed in the district, not to profit from the excess generation, i.e., island operation is incentivized.
\begin{equation}
    \textrm{Net Electricity Demand} = \sum_{t=0}^{n-1}{\textrm{max}(0,Q_t)}
    \label{eqn:cost_function-net_electricity_demand}
\end{equation}

\subsubsection{Ramping} is the difference in net electric consumption at two consecutive timesteps and is defined by Equation \ref{eqn:cost_function-ramping} where $Q_t$ is the net electric consumption at timestep $t$ and $n$ is the total number of timesteps such that $0 \le t < n$.
\begin{equation}
    \textrm{Ramping} = \sum_{t=1}^{n-1}{\lvert Q_t - Q_{t-1} \rvert}
    \label{eqn:cost_function-ramping}
\end{equation}



\subsection{Experimental Design} \label{subsec:experimental_design}
We vary the offline training period and the RBC sequence of operation during offline training to test our hypothesis. The initial 744 (two weeks), 4,344 (six months) and 8,760 (one year) timesteps are used for offline training using either RBC\textsubscript{Basic} or RBC\textsubscript{Optimized} before switching to the SAC or MARLISA agents. Hence, the RL agents considered in totality include:

\begin{enumerate}
    \item SAC\textsubscript{RBC\textsubscript{Basic}}
    \item SAC\textsubscript{RBC\textsubscript{Optimized}}
    \item MARLISA\textsubscript{RBC\textsubscript{Basic}}
    \item MARLISA\textsubscript{RBC\textsubscript{Optimized}}
\end{enumerate}

With these combinations, we study the impact of simpler vs comparatively more complex algorithms (independent SAC vs MARLISA) and the value of less or more detailed domain knowledge (RBC\textsubscript{Basic} vs RBC\textsubscript{Optimized}).

The simulations are run for one epoch, where an epoch is a period of 35,040 timesteps that represent the number of hours in years 2014--2017. We simulate each combination of offline training period and RL agent three times in CityLearn, initialized with different random seeds. The results are averaged over the three runs.

\subsection{Results} \label{sec:results}
\subsubsection{Performance Metrics}
\begin{figure*}
    \centering
    \includegraphics[width=2\columnwidth]{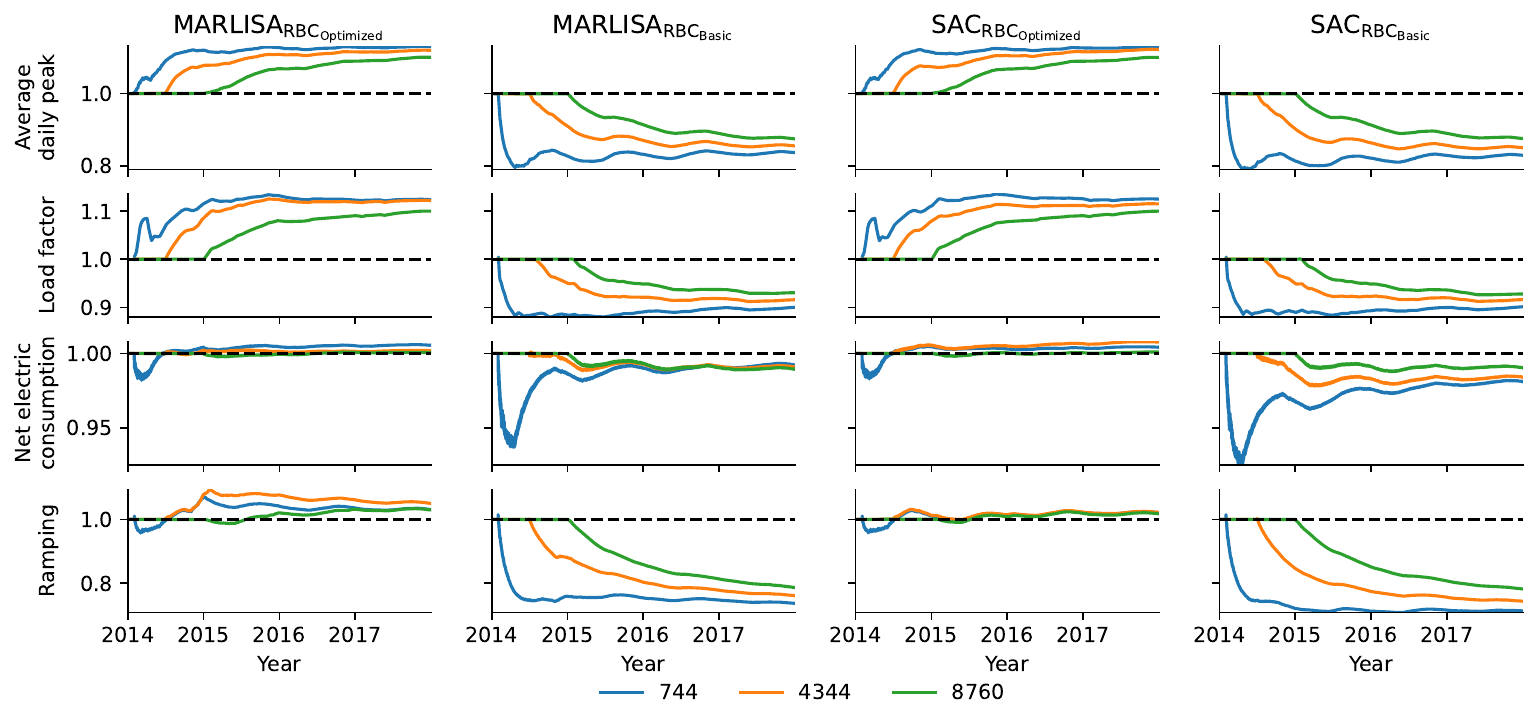}
    \caption{Energy flexibility performance metrics evaluated on results from CityLearn simulations of varied offline training period and RL agents. Offline training periods include 744 (two weeks), 4344 (six months), 8760 (one year) timesteps indicated by the blue, orange, and green lines respectively. The RL agents include SAC\textsubscript{RBC\textsubscript{Basic}}, SAC\textsubscript{RBC\textsubscript{Optimized}}, MARLISA\textsubscript{RBC\textsubscript{Basic}}, MARLISA\textsubscript{RBC\textsubscript{Optimized}}. Each metric is normalized with respect to the RBC used for offline training (dashed black line) which, is indicated in the subscript of the RL agent's name.}
    \label{fig:normalized_cost_function}
\end{figure*}

\cref{fig:normalized_cost_function} shows the performance metrics for the varied offline training periods and RL agents outlined in \cref{subsec:experimental_design} . The metrics are normalized with respect to the RBC used for offline training (dashed black line), where superior and inferior performance of the RL agents is indicated by values less than one and values greater than one, respectively. The detailed domain knowledge of RBC\textsubscript{Optimized} causes superior performance compared to both SAC\textsubscript{RBC\textsubscript{Optimized}} and MARLISA\textsubscript{RBC\textsubscript{Optimized}} agents. Consequently, longer offline training with RBC\textsubscript{Optimized} results in delayed convergence but better performance in the long run. On the other hand, the simplified sequence of operation utilized in RBC\textsubscript{Basic} leads to inferior performance compared to the RL agents, such that longer trained RL agents suffer from poorer performance compared to those trained for a shorter period. The net electric consumption for RBC\textsubscript{Optimized}-trained RL agents is noteworthy, as variations in offline training period show negligible difference in performance. Interestingly, the shortest offline training period of 2 weeks results in an initially large improvement in the net electric consumption metric immediately after the RL agents comes online but within the first year, worsens and approaches the lower performance six-month and one-year trained agents.

Between the SAC and MARLISA RL algorithms, average daily peak and load factor are unaffected by algorithm complexity when the agents are trained using the same RBC. The ramping metric for MARLISA\textsubscript{RBC\textsubscript{Optimized}} shows poor initial performance for shorter offline training periods, but improves over time. In comparison, the SAC\textsubscript{RBC\textsubscript{Optimized}} agents are able to maintain nearly the same ramping performance as RBC\textsubscript{Optimized}.

\subsubsection{District Electricity Consumption}
\begin{figure*}
    \centering
    \includegraphics[width=2\columnwidth]{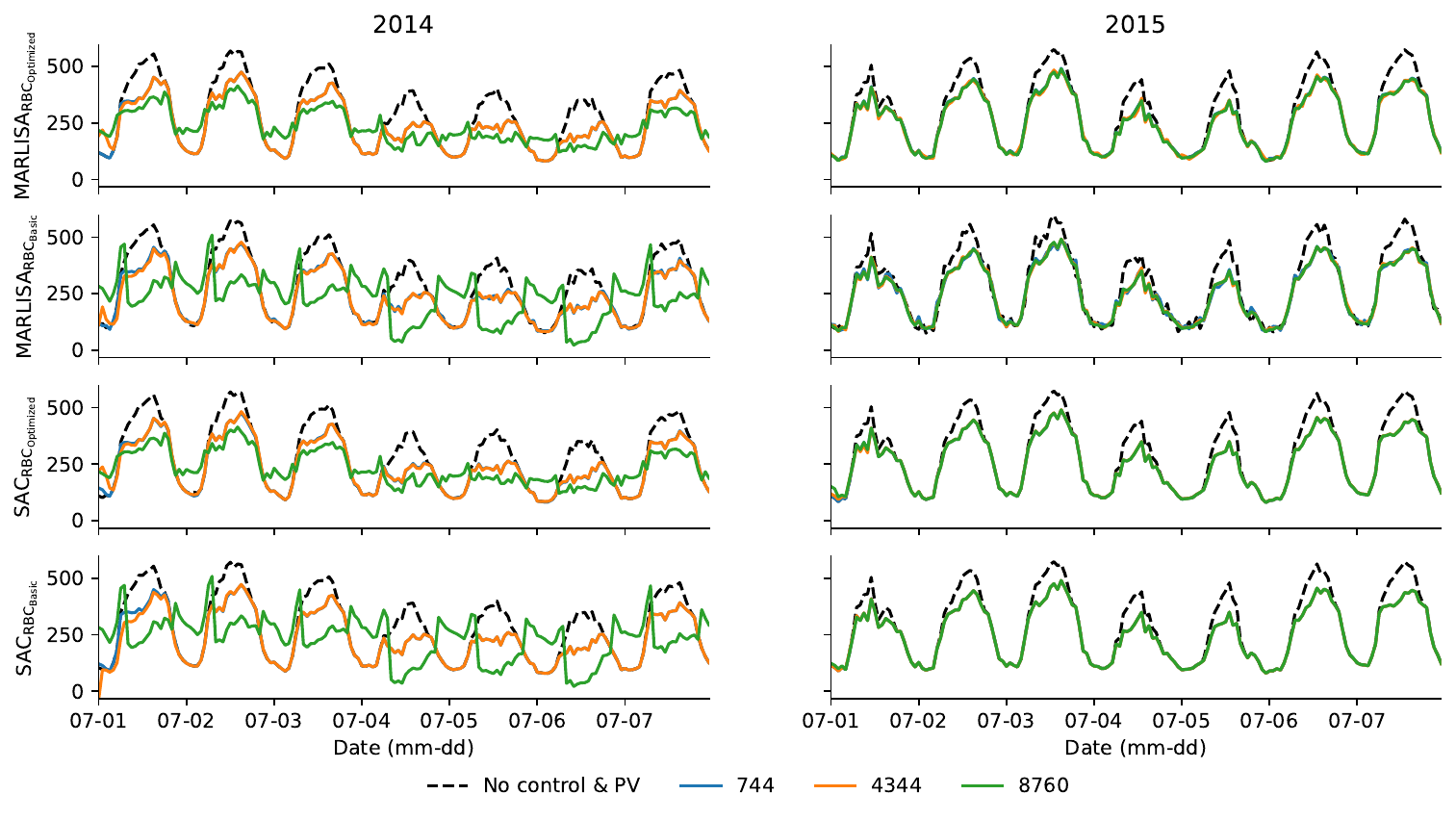}
    \caption{Comparison between electricity consumption without control and PV (dashed black line) and net electricity with RL agent-controlled chilled water DHW and electric storage and, PV at the district level for varied offline training periods. The offline training periods include 744 (two weeks), 4344 (six months), 8760 (one year) timesteps indicated by the blue, orange, and green lines respectively. The RL agents include SAC\textsubscript{RBC\textsubscript{Basic}}, SAC\textsubscript{RBC\textsubscript{Optimized}}, MARLISA\textsubscript{RBC\textsubscript{Basic}}, MARLISA\textsubscript{RBC\textsubscript{Optimized}}. The following seven days after six months of simulation in 2014 are shown (left) and the same time period in the subsequent year of 2015 is shown (right).}
    \label{fig:net_electric_consumption_snapshot}
\end{figure*}

In \cref{fig:net_electric_consumption_snapshot}, we show the district's net electric consumption profile for the four offline trained RL agents, as well its electric consumption without PV installation and energy storage control for a selected period. The 2014 profile is the following seven days after offline training for six months and, the same period is shown in 2015. In 2014, the two-week and six-month trained RL agents are already online while the agents trained for one year are still being trained offline hence, represents net electric consumption under RBC control. For each RL agent, the six-month trained agents behave like the two-week trained agents immediately after coming online and as a result both variations of training period have the same net electric consumption six months into the simulation. For all RL agents in 2014, the one-year training setup still offline has higher net electric consumption early in the morning and late at night, but lower net electric consumption during midday compared to already online scenarios. By the same period in 2015, the net electric consumption profile is almost equal irrespective of RBC domain knowledge, RL algorithm complexity and offline training period. Overall, there is significant energy flexibility in the form of peak shaving provided by solar generation and energy storage systems between late morning and afternoon.

\section{Discussion} \label{sec:discussion}
\subsection{Advanced Building Controllers}
Advanced building controllers are needed to improve upon the industry standard of pre-determined set-points, that do not take into account predictions or allow optimizing the operational sequence~\cite{Wang2020}. Model predictive control (MPC) has been developed in the petrochemical industry in the 1970s and applied across many industries since then~\cite{Morari1999}. MPC requires the development of a mathematical model for the plant to be controlled, which works well for replicable systems (cars, planes). The uniqueness of buildings and their energy systems, and the engineering costs incurred when developing and calibrating a model made it such that despite all advances, MPCs have not been adopted in the building industry~\cite{Kontes2018,privara13}. Reinforcement learning algorithms have been considered to address the shortcomings of MPC by potentially being model-free. However, RL approaches can be more data intensive and more time-consuming compared to MPCs. Comparisons, if even performed, are often biased toward one type of algorithm, and therefore relatively meaningless. The challenges introduced here specifically focus on the breadth of applications rather than on one specific problem. This allows for a fair comparison. Of course, while we argue in the context of reinforcement learning, the challenges can be used for comparisons between algorithm classes.

A promising approach in MARL is centralized training with decentralized execution (CTDE). CTDE assumes that the learning of each agent's policy can depend on the global state (the aggregation of all agents' observation in our case), but during executing, agents work independently. By doing so, it is possible for the agents to cooperate according to some learned heuristics so that during execution they do not need to know what others' observations are. A CTDE version of MARLISA has been found to provide more smooth trajectories compared to the basic MARLISA algorithm~\cite{Glatt2021}. Of course, advances in algorithm complexity must be weighted against data and communication requirements and potential privacy issues.

\subsection{Environment Standardization}
We emphasize the need for standardizing computational environments, such as the COmprehensive Building simulator (COBS)~\cite{Zhang2020a}, Sinergym~\cite{Jimenez-Raboso2021}, BOPTEST~\cite{Blum2021}, the Advanced Controls TestBed (ACTB), or CityLearn~\cite{Vazquez-Canteli2019} using a common interface, e.g., OpenAI Gym~\cite{Brockman2016}, and releasing datasets and implementations open source. This can help spark a development rush similar to the one that the ImageNet dataset sparked for the deep learning community~\cite{JiaDeng2009}. However, in contrast to ImageNet's development, a more in-depth collaboration and exchange between researchers in the built environment and computer science would be beneficial to transfer domain knowledge from buildings to controller design on the one hand and facilitate transitioning theoretical findings of algorithms into practice on the other. Common venues or guest invitation to each other's venues could be established: ACM's BuildSys/e-energy and the ASHRAE/IBPSA communities should explore common pathways for knowledge exchange to ultimately unlock the built environment's potential to reduce greenhouse gas emissions.

\subsection{Offline Learning Challenge (C8)}
Our central hypothesis in addressing \textbf{C8} is that \textit{a longer offline training period results in better performance, since the agents will have more existing knowledge of what ideal actions could resemble by the time they come online}. We find this hypothesis to be true and governed by certain design choices. Our experiments reveal that the sequence of operation utilized in the RBC used for offline training determines the performance of the RL agents when evaluated on a set of energy flexibility metrics. Longer offline learning from an optimized RBC will lead to slower convergence upon coming online, but superior energy flexibility in the long run. RL agents that learn from a simplified RBC risk poorer performance as the offline learning period increases.

The optimized RBC is able to significantly outperform the RL controllers in reducing the district average daily peak, load factor and ramping. This shows significant energy flexibility potential from improving existing RBCs in practice over installation of more complex controllers. Nevertheless, RBC controlled systems are unable to respond to perturbations in the control environment (\textbf{C5}), an ability RL controllers possess, which may affect the overall performance of the controller in satisfying the control objective. We hope to address \textbf{C5} in our future work.

We do not observe any significant differences between the performance of the SAC and MARLISA RL algorithms when evaluated on the 4 performance metrics. This suggests that the simpler SAC algorithm is sufficient and the added complexity and cost of information sharing amongst agents could be avoided.

Our experiments show negligible difference in net electricity consumption irrespective of offline learning period, RBC sequence of operation and RL algorithm complexity. We provide an explanation in the context of the RBC design. Both RBC\textsubscript{Basic} and RBC\textsubscript{Optimized} are designed to charge the storage systems at night and early in the morning to take advantage of higher heat pump COP. Their logic can also be beneficial in a residential DR program that incentivizes electricity consumption during periods of lower demand. However, in the absence of such DR setup in the simulation environment, this design is most beneficial to the chilled water storage whose energy is delivered by a heat pump. The DHW and electrical storage charging demands are directly met by the grid and offset by available solar generation. Solar generation is intermittently available during the day, hence, these storage devices could potentially benefit from 'free' charging during the RBC's hours of discharge control action.

\section{Conclusion} \label{sec:conclusion}
We have introduced a set of challenges to study real world grid-interactive buildings. While there are many research challenges that remain in this realm, we highlight the need for an organized move forward of the community in addressing both fundamental computational challenges, but in a way that applies to the larger problems in the built environment. As an example, we studied the off-line learning challenge (C8) for two levels of domain knowledge, RL algorithm complexity and five performance metrics. It is not our intention to imply that the list above is an exhaustive list of challenges. Rather, by highlighting typical real world problems, our aim is to inspire researchers to define and share their environments and the problems they are addressing with these challenges as a standard framework.

\balance
\bibliographystyle{ACM-Reference-Format}
\bibliography{references.bib}


\begin{thebibliography}{38}


\ifx \showCODEN    \undefined \def \showCODEN     #1{\unskip}     \fi
\ifx \showDOI      \undefined \def \showDOI       #1{#1}\fi
\ifx \showISBNx    \undefined \def \showISBNx     #1{\unskip}     \fi
\ifx \showISBNxiii \undefined \def \showISBNxiii  #1{\unskip}     \fi
\ifx \showISSN     \undefined \def \showISSN      #1{\unskip}     \fi
\ifx \showLCCN     \undefined \def \showLCCN      #1{\unskip}     \fi
\ifx \shownote     \undefined \def \shownote      #1{#1}          \fi
\ifx \showarticletitle \undefined \def \showarticletitle #1{#1}   \fi
\ifx \showURL      \undefined \def \showURL       {\relax}        \fi
\providecommand\bibfield[2]{#2}
\providecommand\bibinfo[2]{#2}
\providecommand\natexlab[1]{#1}
\providecommand\showeprint[2][]{arXiv:#2}

\bibitem[\protect\citeauthoryear{Blum, Arroyo, Huang, Drgoňa, Jorissen,
  Walnum, Chen, Benne, Vrabie, Wetter, and Helsen}{Blum et~al\mbox{.}}{2021}]%
        {Blum2021}
\bibfield{author}{\bibinfo{person}{David Blum}, \bibinfo{person}{Javier
  Arroyo}, \bibinfo{person}{Sen Huang}, \bibinfo{person}{Ján Drgoňa},
  \bibinfo{person}{Filip Jorissen}, \bibinfo{person}{Harald~Taxt Walnum},
  \bibinfo{person}{Yan Chen}, \bibinfo{person}{Kyle Benne},
  \bibinfo{person}{Draguna Vrabie}, \bibinfo{person}{Michael Wetter}, {and}
  \bibinfo{person}{Lieve Helsen}.} \bibinfo{year}{2021}\natexlab{}.
\newblock \showarticletitle{{Building optimization testing framework (BOPTEST)
  for simulation-based benchmarking of control strategies in buildings}}.
\newblock \bibinfo{journal}{\emph{Journal of Building Performance Simulation}}
  \bibinfo{volume}{14}, \bibinfo{number}{5} (\bibinfo{date}{9}
  \bibinfo{year}{2021}), \bibinfo{pages}{586--610}.
\newblock
\showISSN{1940-1493}
\urldef\tempurl%
\url{https://doi.org/10.1080/19401493.2021.1986574}
\showDOI{\tempurl}


\bibitem[\protect\citeauthoryear{Brockman, Cheung, Pettersson, Schneider,
  Schulman, Tang, and Zaremba}{Brockman et~al\mbox{.}}{2016}]%
        {Brockman2016}
\bibfield{author}{\bibinfo{person}{Greg Brockman}, \bibinfo{person}{Vicki
  Cheung}, \bibinfo{person}{Ludwig Pettersson}, \bibinfo{person}{Jonas
  Schneider}, \bibinfo{person}{John Schulman}, \bibinfo{person}{Jie Tang},
  {and} \bibinfo{person}{Wojciech Zaremba}.} \bibinfo{year}{2016}\natexlab{}.
\newblock \showarticletitle{{OpenAI Gym}}.
\newblock \bibinfo{journal}{\emph{arxiv preprint: 1606.01540}}
  (\bibinfo{date}{6} \bibinfo{year}{2016}).
\newblock
\urldef\tempurl%
\url{http://arxiv.org/abs/1606.01540}
\showURL{%
\tempurl}


\bibitem[\protect\citeauthoryear{Bruninx, Patteeuw, Delarue, Helsen, and
  D'Haeseleer}{Bruninx et~al\mbox{.}}{2013}]%
        {Bruninx2013Short-termSimulations}
\bibfield{author}{\bibinfo{person}{Kenneth Bruninx}, \bibinfo{person}{Dieter
  Patteeuw}, \bibinfo{person}{Erik Delarue}, \bibinfo{person}{Lieve Helsen},
  {and} \bibinfo{person}{William D'Haeseleer}.}
  \bibinfo{year}{2013}\natexlab{}.
\newblock \showarticletitle{{Short-term demand response of flexible electric
  heating systems: The need for integrated simulations}}.
\newblock \bibinfo{journal}{\emph{International Conference on the European
  Energy Market, EEM}} \bibinfo{number}{May} (\bibinfo{year}{2013}),
  \bibinfo{pages}{28--30}.
\newblock
\showISBNx{9781479920082}
\showISSN{21654077}
\urldef\tempurl%
\url{https://doi.org/10.1109/EEM.2013.6607333}
\showDOI{\tempurl}


\bibitem[\protect\citeauthoryear{Chourabi, Nam, Walker, Gil-Garcia, Mellouli,
  Nahon, Pardo, and Scholl}{Chourabi et~al\mbox{.}}{2011}]%
        {Chourabi2011UnderstandingFramework}
\bibfield{author}{\bibinfo{person}{Hafedh Chourabi}, \bibinfo{person}{Taewoo
  Nam}, \bibinfo{person}{Shawn Walker}, \bibinfo{person}{J.~Ramon Gil-Garcia},
  \bibinfo{person}{Sehl Mellouli}, \bibinfo{person}{Karine Nahon},
  \bibinfo{person}{Theresa~A. Pardo}, {and} \bibinfo{person}{Hans~Jochen
  Scholl}.} \bibinfo{year}{2011}\natexlab{}.
\newblock \showarticletitle{{Understanding smart cities: An integrative
  framework}}.
\newblock \bibinfo{journal}{\emph{Proceedings of the Annual Hawaii
  International Conference on System Sciences}} (\bibinfo{year}{2011}),
  \bibinfo{pages}{2289--2297}.
\newblock
\showISBNx{9780769545257}
\showISSN{15301605}
\urldef\tempurl%
\url{https://doi.org/10.1109/HICSS.2012.615}
\showDOI{\tempurl}


\bibitem[\protect\citeauthoryear{Deltetto, Coraci, Pinto, Piscitelli, and
  Capozzoli}{Deltetto et~al\mbox{.}}{2021}]%
        {Deltetto2021}
\bibfield{author}{\bibinfo{person}{Davide Deltetto}, \bibinfo{person}{Davide
  Coraci}, \bibinfo{person}{Giuseppe Pinto}, \bibinfo{person}{Marco~Savino
  Piscitelli}, {and} \bibinfo{person}{Alfonso Capozzoli}.}
  \bibinfo{year}{2021}\natexlab{}.
\newblock \showarticletitle{{Exploring the potentialities of deep reinforcement
  learning for incentive-based demand response in a cluster of small commercial
  buildings}}.
\newblock \bibinfo{journal}{\emph{Energies}} \bibinfo{volume}{14},
  \bibinfo{number}{10} (\bibinfo{year}{2021}).
\newblock
\showISSN{19961073}
\urldef\tempurl%
\url{https://doi.org/10.3390/en14102933}
\showDOI{\tempurl}


\bibitem[\protect\citeauthoryear{{Department of Energy}}{{Department of
  Energy}}{2021}]%
        {DOE2021}
\bibfield{author}{\bibinfo{person}{{Department of Energy}}.}
  \bibinfo{year}{2021}\natexlab{}.
\newblock \bibinfo{booktitle}{\emph{{A National Roadmap for Grid-Interactive
  Efficient Buildings}}}.
\newblock \bibinfo{type}{{T}echnical {R}eport}.
  \bibinfo{institution}{Department of Energy}. \bibinfo{pages}{166} pages.
\newblock


\bibitem[\protect\citeauthoryear{Dhamankar, Vazquez-Canteli, and
  Nagy}{Dhamankar et~al\mbox{.}}{2020}]%
        {Dhamankar2020}
\bibfield{author}{\bibinfo{person}{Gauraang Dhamankar},
  \bibinfo{person}{Jose~R. Vazquez-Canteli}, {and} \bibinfo{person}{Zoltan
  Nagy}.} \bibinfo{year}{2020}\natexlab{}.
\newblock \showarticletitle{{Benchmarking Multi-Agent Deep Reinforcement
  Learning Algorithms on a Building Energy Demand Coordination Task}}.
\newblock \bibinfo{journal}{\emph{RLEM 2020 - Proceedings of the 1st
  International Workshop on Reinforcement Learning for Energy Management in
  Buildings and Cities}} (\bibinfo{year}{2020}), \bibinfo{pages}{15--19}.
\newblock
\showISBNx{9781450381932}
\urldef\tempurl%
\url{https://doi.org/10.1145/3427773.3427870}
\showDOI{\tempurl}


\bibitem[\protect\citeauthoryear{Drgoňa, Arroyo, Cupeiro~Figueroa, Blum,
  Arendt, Kim, Oll{\'{e}}, Oravec, Wetter, Vrabie, and Helsen}{Drgoňa
  et~al\mbox{.}}{2020}]%
        {Drgona2020}
\bibfield{author}{\bibinfo{person}{Ján Drgoňa}, \bibinfo{person}{Javier
  Arroyo}, \bibinfo{person}{Iago Cupeiro~Figueroa}, \bibinfo{person}{David
  Blum}, \bibinfo{person}{Krzysztof Arendt}, \bibinfo{person}{Donghun Kim},
  \bibinfo{person}{Enric~Perarnau Oll{\'{e}}}, \bibinfo{person}{Juraj Oravec},
  \bibinfo{person}{Michael Wetter}, \bibinfo{person}{Draguna~L. Vrabie}, {and}
  \bibinfo{person}{Lieve Helsen}.} \bibinfo{year}{2020}\natexlab{}.
\newblock \showarticletitle{{All you need to know about model predictive
  control for buildings}}.
\newblock \bibinfo{journal}{\emph{Annual Reviews in Control}}
  \bibinfo{volume}{50}, \bibinfo{number}{May} (\bibinfo{year}{2020}),
  \bibinfo{pages}{190--232}.
\newblock
\showISSN{13675788}
\urldef\tempurl%
\url{https://doi.org/10.1016/j.arcontrol.2020.09.001}
\showDOI{\tempurl}


\bibitem[\protect\citeauthoryear{Drgoňa, Tuor, Chandan, and Vrabie}{Drgoňa
  et~al\mbox{.}}{2021}]%
        {Drgona2021}
\bibfield{author}{\bibinfo{person}{Ján Drgoňa}, \bibinfo{person}{Aaron~R.
  Tuor}, \bibinfo{person}{Vikas Chandan}, {and} \bibinfo{person}{Draguna~L.
  Vrabie}.} \bibinfo{year}{2021}\natexlab{}.
\newblock \showarticletitle{{Physics-constrained deep learning of multi-zone
  building thermal dynamics}}.
\newblock \bibinfo{journal}{\emph{Energy and Buildings}}  \bibinfo{volume}{243}
  (\bibinfo{year}{2021}).
\newblock
\showISSN{03787788}
\urldef\tempurl%
\url{https://doi.org/10.1016/j.enbuild.2021.110992}
\showDOI{\tempurl}


\bibitem[\protect\citeauthoryear{Dulac-Arnold, Levine, Mankowitz, Li, Paduraru,
  Gowal, and Hester}{Dulac-Arnold et~al\mbox{.}}{2021}]%
        {Dulac-Arnold2021}
\bibfield{author}{\bibinfo{person}{Gabriel Dulac-Arnold}, \bibinfo{person}{Nir
  Levine}, \bibinfo{person}{Daniel~J. Mankowitz}, \bibinfo{person}{Jerry Li},
  \bibinfo{person}{Cosmin Paduraru}, \bibinfo{person}{Sven Gowal}, {and}
  \bibinfo{person}{Todd Hester}.} \bibinfo{year}{2021}\natexlab{}.
\newblock \bibinfo{booktitle}{\emph{{Challenges of real-world reinforcement
  learning: definitions, benchmarks and analysis}}}.
\newblock Number 0123456789. \bibinfo{publisher}{Springer US}.
\newblock
\showISBNx{0123456789}
\showISSN{15730565}
\urldef\tempurl%
\url{https://doi.org/10.1007/s10994-021-05961-4}
\showDOI{\tempurl}


\bibitem[\protect\citeauthoryear{Dupont, Dietrich, De~Jonghe, Ramos, and
  Belmans}{Dupont et~al\mbox{.}}{2014}]%
        {Dupont2014ImpactStudy}
\bibfield{author}{\bibinfo{person}{B. Dupont}, \bibinfo{person}{K. Dietrich},
  \bibinfo{person}{C. De~Jonghe}, \bibinfo{person}{A. Ramos}, {and}
  \bibinfo{person}{R. Belmans}.} \bibinfo{year}{2014}\natexlab{}.
\newblock \showarticletitle{{Impact of residential demand response on power
  system operation: A Belgian case study}}.
\newblock \bibinfo{journal}{\emph{Applied Energy}}  \bibinfo{volume}{122}
  (\bibinfo{year}{2014}), \bibinfo{pages}{1--10}.
\newblock
\showISBNx{0306-2619}
\showISSN{03062619}
\urldef\tempurl%
\url{https://doi.org/10.1016/j.apenergy.2014.02.022}
\showDOI{\tempurl}


\bibitem[\protect\citeauthoryear{{Github}}{{Github}}{[n.d.]}]%
        {citylearngithub}
\bibfield{author}{\bibinfo{person}{{Github}}.}
  \bibinfo{year}{[n.d.]}\natexlab{}.
\newblock
  \bibinfo{title}{{https://github.com/intelligent-environments-lab/CityLearn}}.
\newblock
\newblock


\bibitem[\protect\citeauthoryear{Glatt, Silva, Soper, Dawson, Rusu, and
  Goldhahn}{Glatt et~al\mbox{.}}{2021}]%
        {Glatt2021}
\bibfield{author}{\bibinfo{person}{Ruben Glatt}, \bibinfo{person}{Felipe
  Leno~da Silva}, \bibinfo{person}{Braden Soper}, \bibinfo{person}{William~A.
  Dawson}, \bibinfo{person}{Edward Rusu}, {and} \bibinfo{person}{Ryan~A.
  Goldhahn}.} \bibinfo{year}{2021}\natexlab{}.
\newblock \showarticletitle{{Collaborative energy demand response with
  decentralized actor and centralized critic}}. In
  \bibinfo{booktitle}{\emph{Proceedings of the 8th ACM International Conference
  on Systems for Energy-Efficient Buildings, Cities, and Transportation}}.
  \bibinfo{publisher}{ACM}, \bibinfo{address}{New York, NY, USA},
  \bibinfo{pages}{333--337}.
\newblock
\showISBNx{9781450391146}
\urldef\tempurl%
\url{https://doi.org/10.1145/3486611.3488732}
\showDOI{\tempurl}


\bibitem[\protect\citeauthoryear{Haarnoja, Zhou, Abbeel, and Levine}{Haarnoja
  et~al\mbox{.}}{2018a}]%
        {Haarnoja2018}
\bibfield{author}{\bibinfo{person}{Tuomas Haarnoja}, \bibinfo{person}{Aurick
  Zhou}, \bibinfo{person}{Pieter Abbeel}, {and} \bibinfo{person}{Sergey
  Levine}.} \bibinfo{year}{2018}\natexlab{a}.
\newblock \showarticletitle{{Soft Actor-Critic: Off-Policy Maximum Entropy Deep
  Reinforcement Learning with a Stochastic Actor}}. In
  \bibinfo{booktitle}{\emph{ICML}}.
\newblock
\urldef\tempurl%
\url{http://arxiv.org/abs/1801.01290}
\showURL{%
\tempurl}


\bibitem[\protect\citeauthoryear{Haarnoja, Zhou, Hartikainen, Tucker, Ha, Tan,
  Kumar, Zhu, Gupta, Abbeel, and Levine}{Haarnoja et~al\mbox{.}}{2018b}]%
        {Haarnoja2018SoftAA}
\bibfield{author}{\bibinfo{person}{Tuomas Haarnoja}, \bibinfo{person}{Aurick
  Zhou}, \bibinfo{person}{Kristian Hartikainen}, \bibinfo{person}{G. Tucker},
  \bibinfo{person}{Sehoon Ha}, \bibinfo{person}{Jie Tan},
  \bibinfo{person}{Vikash Kumar}, \bibinfo{person}{Henry Zhu},
  \bibinfo{person}{Abhishek Gupta}, \bibinfo{person}{P. Abbeel}, {and}
  \bibinfo{person}{Sergey Levine}.} \bibinfo{year}{2018}\natexlab{b}.
\newblock \showarticletitle{Soft Actor-Critic Algorithms and Applications}.
\newblock \bibinfo{journal}{\emph{ArXiv}}  \bibinfo{volume}{abs/1812.05905}
  (\bibinfo{year}{2018}).
\newblock


\bibitem[\protect\citeauthoryear{{Jia Deng}, {Wei Dong}, Socher, {Li-Jia Li},
  {Kai Li}, and {Li Fei-Fei}}{{Jia Deng} et~al\mbox{.}}{2009}]%
        {JiaDeng2009}
\bibfield{author}{\bibinfo{person}{{Jia Deng}}, \bibinfo{person}{{Wei Dong}},
  \bibinfo{person}{R. Socher}, \bibinfo{person}{{Li-Jia Li}},
  \bibinfo{person}{{Kai Li}}, {and} \bibinfo{person}{{Li Fei-Fei}}.}
  \bibinfo{year}{2009}\natexlab{}.
\newblock \showarticletitle{{ImageNet: A large-scale hierarchical image
  database}}. In \bibinfo{booktitle}{\emph{2009 IEEE Conference on Computer
  Vision and Pattern Recognition}}. \bibinfo{publisher}{IEEE},
  \bibinfo{pages}{248--255}.
\newblock
\urldef\tempurl%
\url{https://doi.org/10.1109/CVPRW.2009.5206848}
\showDOI{\tempurl}


\bibitem[\protect\citeauthoryear{Jim{\'{e}}nez-Raboso, Campoy-Nieves,
  Manjavacas-Lucas, G{\'{o}}mez-Romero, and Molina-Solana}{Jim{\'{e}}nez-Raboso
  et~al\mbox{.}}{2021}]%
        {Jimenez-Raboso2021}
\bibfield{author}{\bibinfo{person}{Javier Jim{\'{e}}nez-Raboso},
  \bibinfo{person}{Alejandro Campoy-Nieves}, \bibinfo{person}{Antonio
  Manjavacas-Lucas}, \bibinfo{person}{Juan G{\'{o}}mez-Romero}, {and}
  \bibinfo{person}{Miguel Molina-Solana}.} \bibinfo{year}{2021}\natexlab{}.
\newblock \showarticletitle{{Sinergym: a building simulation and control
  framework for training reinforcement learning agents}}. In
  \bibinfo{booktitle}{\emph{Proceedings of the 8th ACM International Conference
  on Systems for Energy-Efficient Buildings, Cities, and Transportation}}.
  \bibinfo{publisher}{ACM}, \bibinfo{address}{New York, NY, USA},
  \bibinfo{pages}{319--323}.
\newblock
\showISBNx{9781450391146}
\urldef\tempurl%
\url{https://doi.org/10.1145/3486611.3488729}
\showDOI{\tempurl}


\bibitem[\protect\citeauthoryear{Kathirgamanathan, Twardowski, Mangina, and
  Finn}{Kathirgamanathan et~al\mbox{.}}{2020}]%
        {Kathirgamanathan2020}
\bibfield{author}{\bibinfo{person}{Anjukan Kathirgamanathan},
  \bibinfo{person}{Kacper Twardowski}, \bibinfo{person}{Eleni Mangina}, {and}
  \bibinfo{person}{Donal~P. Finn}.} \bibinfo{year}{2020}\natexlab{}.
\newblock \showarticletitle{{A Centralised Soft Actor Critic Deep Reinforcement
  Learning Approach to District Demand Side Management through CityLearn}}. In
  \bibinfo{booktitle}{\emph{Proceedings of the 1st International Workshop on
  Reinforcement Learning for Energy Management in Buildings {\&} Cities}}.
  \bibinfo{publisher}{ACM}, \bibinfo{address}{New York, NY, USA},
  \bibinfo{pages}{11--14}.
\newblock
\showISBNx{9781450381932}
\urldef\tempurl%
\url{https://doi.org/10.1145/3427773.3427869}
\showDOI{\tempurl}


\bibitem[\protect\citeauthoryear{Kontes, Giannakis, S{\'{a}}nchez,
  de~Agustin-Camacho, Romero-Amorrortu, Panagiotidou, Rovas, Steiger,
  Mutschler, and Gruen}{Kontes et~al\mbox{.}}{2018}]%
        {Kontes2018}
\bibfield{author}{\bibinfo{person}{Georgios~D. Kontes},
  \bibinfo{person}{Georgios~I. Giannakis}, \bibinfo{person}{Víctor
  S{\'{a}}nchez}, \bibinfo{person}{Pablo de Agustin-Camacho},
  \bibinfo{person}{Ander Romero-Amorrortu}, \bibinfo{person}{Natalia
  Panagiotidou}, \bibinfo{person}{Dimitrios~V. Rovas}, \bibinfo{person}{Simone
  Steiger}, \bibinfo{person}{Christopher Mutschler}, {and}
  \bibinfo{person}{Gunnar Gruen}.} \bibinfo{year}{2018}\natexlab{}.
\newblock \showarticletitle{{Simulation-based evaluation and optimization of
  control strategies in buildings}}.
\newblock \bibinfo{journal}{\emph{Energies}} \bibinfo{volume}{11},
  \bibinfo{number}{12} (\bibinfo{year}{2018}), \bibinfo{pages}{1--23}.
\newblock
\showISBNx{4991158061325}
\showISSN{19961073}
\urldef\tempurl%
\url{https://doi.org/10.3390/en11123376}
\showDOI{\tempurl}


\bibitem[\protect\citeauthoryear{Leibowicz, Lanham, Brozynski, Vazquez-Canteli,
  Castejon, and Nagy}{Leibowicz et~al\mbox{.}}{2018}]%
        {Lanham2018}
\bibfield{author}{\bibinfo{person}{Benjamin~D. Leibowicz},
  \bibinfo{person}{Christopher~M. Lanham}, \bibinfo{person}{Max~T. Brozynski},
  \bibinfo{person}{Jose~R. Vazquez-Canteli}, \bibinfo{person}{Nicolas~Castillo
  Castejon}, {and} \bibinfo{person}{Zoltan Nagy}.}
  \bibinfo{year}{2018}\natexlab{}.
\newblock \showarticletitle{{Optimal decarbonization pathways for urban
  residential building energy services}}.
\newblock \bibinfo{journal}{\emph{Applied Energy}} \bibinfo{volume}{230},
  \bibinfo{number}{May} (\bibinfo{date}{11} \bibinfo{year}{2018}),
  \bibinfo{pages}{1311--1325}.
\newblock
\urldef\tempurl%
\url{https://doi.org/10.1016/j.apenergy.2018.09.046}
\showDOI{\tempurl}


\bibitem[\protect\citeauthoryear{Lucon and {\"{U}}rge-Vorsatz}{Lucon and
  {\"{U}}rge-Vorsatz}{2014}]%
        {ipcc14}
\bibfield{author}{\bibinfo{person}{O. Lucon} {and} \bibinfo{person}{D.
  {\"{U}}rge-Vorsatz}.} \bibinfo{year}{2014}\natexlab{}.
\newblock \showarticletitle{{Fifth Assessment Report, Mitigation of Climate
  Change}}.
\newblock \bibinfo{journal}{\emph{Intergovernmental Panel on Climate Change}}
  (\bibinfo{year}{2014}), \bibinfo{pages}{674--738}.
\newblock


\bibitem[\protect\citeauthoryear{Mohagheghi, Stoupis, Wang, and Li}{Mohagheghi
  et~al\mbox{.}}{2010}]%
        {Mohagheghi2010DemandSystem}
\bibfield{author}{\bibinfo{person}{S Mohagheghi}, \bibinfo{person}{J Stoupis},
  \bibinfo{person}{Z Wang}, {and} \bibinfo{person}{Z Li}.}
  \bibinfo{year}{2010}\natexlab{}.
\newblock \showarticletitle{{Demand Response Architecture-Integration into the
  Distribution Management System}}.
\newblock \bibinfo{journal}{\emph{SmartGridComm}} (\bibinfo{year}{2010}),
  \bibinfo{pages}{501--506}.
\newblock
\showISBNx{9781424465118}


\bibitem[\protect\citeauthoryear{Morari and H.~Lee}{Morari and H.~Lee}{1999}]%
        {Morari1999}
\bibfield{author}{\bibinfo{person}{Manfred Morari} {and} \bibinfo{person}{Jay
  H.~Lee}.} \bibinfo{year}{1999}\natexlab{}.
\newblock \showarticletitle{{Model predictive control: Past, present and
  future}}.
\newblock \bibinfo{journal}{\emph{Computers and Chemical Engineering}}
  \bibinfo{volume}{23}, \bibinfo{number}{4-5} (\bibinfo{year}{1999}),
  \bibinfo{pages}{667--682}.
\newblock
\showISSN{00981354}
\urldef\tempurl%
\url{https://doi.org/10.1016/S0098-1354(98)00301-9}
\showDOI{\tempurl}


\bibitem[\protect\citeauthoryear{Nagy, Park, and Vazquez-Canteli}{Nagy
  et~al\mbox{.}}{2018}]%
        {Nagy2018}
\bibfield{author}{\bibinfo{person}{Zoltan Nagy}, \bibinfo{person}{June~Young
  Park}, {and} \bibinfo{person}{Jose Vazquez-Canteli}.}
  \bibinfo{year}{2018}\natexlab{}.
\newblock \showarticletitle{{Reinforcement learning for intelligent
  environments: A Tutorial}}.
\newblock In \bibinfo{booktitle}{\emph{Handbook of Sustainable and Resilient
  Infrastructure} (\bibinfo{edition}{1} ed.)},
  \bibfield{editor}{\bibinfo{person}{Paolo Gardoni}} (Ed.).
  \bibinfo{publisher}{Routledge}, Chapter~37.
\newblock


\bibitem[\protect\citeauthoryear{Nagy, V{\'{a}}zquez-Canteli, Dey, and
  Henze}{Nagy et~al\mbox{.}}{2021}]%
        {Nagy2021}
\bibfield{author}{\bibinfo{person}{Zoltan Nagy}, \bibinfo{person}{José~R.
  V{\'{a}}zquez-Canteli}, \bibinfo{person}{Sourav Dey}, {and}
  \bibinfo{person}{Gregor Henze}.} \bibinfo{year}{2021}\natexlab{}.
\newblock \showarticletitle{{The citylearn challenge 2021}}. In
  \bibinfo{booktitle}{\emph{Proceedings of the 8th ACM International Conference
  on Systems for Energy-Efficient Buildings, Cities, and Transportation}}.
  \bibinfo{publisher}{ACM}, \bibinfo{address}{New York, NY, USA},
  \bibinfo{pages}{218--219}.
\newblock
\showISBNx{9781450391146}
\urldef\tempurl%
\url{https://doi.org/10.1145/3486611.3492226}
\showDOI{\tempurl}


\bibitem[\protect\citeauthoryear{Pinto, Piscitelli, V{\'{a}}zquez-Canteli,
  Nagy, and Capozzoli}{Pinto et~al\mbox{.}}{2021}]%
        {Pinto2021}
\bibfield{author}{\bibinfo{person}{Giuseppe Pinto},
  \bibinfo{person}{Marco~Savino Piscitelli}, \bibinfo{person}{José~Ramón
  V{\'{a}}zquez-Canteli}, \bibinfo{person}{Zoltán Nagy}, {and}
  \bibinfo{person}{Alfonso Capozzoli}.} \bibinfo{year}{2021}\natexlab{}.
\newblock \showarticletitle{{Coordinated energy management for a cluster of
  buildings through deep reinforcement learning}}.
\newblock \bibinfo{journal}{\emph{Energy}}  \bibinfo{volume}{229}
  (\bibinfo{year}{2021}).
\newblock
\showISSN{03605442}
\urldef\tempurl%
\url{https://doi.org/10.1016/j.energy.2021.120725}
\showDOI{\tempurl}


\bibitem[\protect\citeauthoryear{Pr{\'{i}}vara, Cigler, V{\'{a}}ňa,
  Oldewurtel, Sagerschnig, and {\v{Z}}{\'{a}}{\v{c}}ekov{\'{a}}}{Pr{\'{i}}vara
  et~al\mbox{.}}{2013}]%
        {privara13}
\bibfield{author}{\bibinfo{person}{Samuel Pr{\'{i}}vara},
  \bibinfo{person}{Jiří Cigler}, \bibinfo{person}{Zdeněk V{\'{a}}ňa},
  \bibinfo{person}{Frauke Oldewurtel}, \bibinfo{person}{Carina Sagerschnig},
  {and} \bibinfo{person}{Eva {\v{Z}}{\'{a}}{\v{c}}ekov{\'{a}}}.}
  \bibinfo{year}{2013}\natexlab{}.
\newblock \showarticletitle{{Building modeling as a crucial part for building
  predictive control}}.
\newblock \bibinfo{journal}{\emph{Energy and Buildings}}  \bibinfo{volume}{56}
  (\bibinfo{year}{2013}), \bibinfo{pages}{8--22}.
\newblock
\showISBNx{0378-7788}
\showISSN{03787788}
\urldef\tempurl%
\url{https://doi.org/10.1016/j.enbuild.2012.10.024}
\showDOI{\tempurl}


\bibitem[\protect\citeauthoryear{Qin, Gao, Zhang, Xu, Huang, Li, Zhang, and
  Yu}{Qin et~al\mbox{.}}{2021}]%
        {Qin2021}
\bibfield{author}{\bibinfo{person}{Rongjun Qin}, \bibinfo{person}{Songyi Gao},
  \bibinfo{person}{Xingyuan Zhang}, \bibinfo{person}{Zhen Xu},
  \bibinfo{person}{Shengkai Huang}, \bibinfo{person}{Zewen Li},
  \bibinfo{person}{Weinan Zhang}, {and} \bibinfo{person}{Yang Yu}.}
  \bibinfo{year}{2021}\natexlab{}.
\newblock \showarticletitle{{NeoRL: A Near Real-World Benchmark for Offline
  Reinforcement Learning}}.
\newblock  (\bibinfo{date}{2} \bibinfo{year}{2021}).
\newblock
\urldef\tempurl%
\url{http://arxiv.org/abs/2102.00714}
\showURL{%
\tempurl}


\bibitem[\protect\citeauthoryear{{Richard S. Sutton and Andrew G.
  Barto}}{{Richard S. Sutton and Andrew G. Barto}}{2018}]%
        {RichardS.SuttonandAndrewG.Barto2018a}
\bibfield{author}{\bibinfo{person}{{Richard S. Sutton and Andrew G. Barto}}.}
  \bibinfo{year}{2018}\natexlab{}.
\newblock \bibinfo{booktitle}{\emph{{Reinforcement Learning, Second Edition An
  Introduction}}}.
\newblock 550 pages.
\newblock
\showISBNx{9780262039246}


\bibitem[\protect\citeauthoryear{Siano}{Siano}{2014}]%
        {Siano2014}
\bibfield{author}{\bibinfo{person}{Pierluigi Siano}.}
  \bibinfo{year}{2014}\natexlab{}.
\newblock \showarticletitle{{Demand response and smart grids - A survey}}.
\newblock \bibinfo{journal}{\emph{Renewable and Sustainable Energy Reviews}}
  \bibinfo{volume}{30} (\bibinfo{year}{2014}), \bibinfo{pages}{461--478}.
\newblock
\showISBNx{13640321 (ISSN)}
\showISSN{13640321}
\urldef\tempurl%
\url{https://doi.org/10.1016/j.rser.2013.10.022}
\showDOI{\tempurl}


\bibitem[\protect\citeauthoryear{V{\'{a}}zquez-Canteli, K{\"{a}}mpf, Henze, and
  Nagy}{V{\'{a}}zquez-Canteli et~al\mbox{.}}{2019}]%
        {Vazquez-Canteli2019}
\bibfield{author}{\bibinfo{person}{J.R. José~R. V{\'{a}}zquez-Canteli},
  \bibinfo{person}{Jérôme K{\"{a}}mpf}, \bibinfo{person}{Gregor Henze}, {and}
  \bibinfo{person}{Zoltan Nagy}.} \bibinfo{year}{2019}\natexlab{}.
\newblock \showarticletitle{{CityLearn v1.0: An OpenAI gym environment for
  demand response with deep reinforcement learning}}.
\newblock \bibinfo{journal}{\emph{BuildSys 2019 - Proceedings of the 6th ACM
  International Conference on Systems for Energy-Efficient Buildings, Cities,
  and Transportation}} (\bibinfo{year}{2019}), \bibinfo{pages}{356--357}.
\newblock
\showISBNx{9781450370059}
\urldef\tempurl%
\url{https://doi.org/10.1145/3360322.3360998}
\showDOI{\tempurl}


\bibitem[\protect\citeauthoryear{Vazquez-Canteli, Dey, Henze, and
  Nagy}{Vazquez-Canteli et~al\mbox{.}}{2020a}]%
        {citylearnArxiv}
\bibfield{author}{\bibinfo{person}{Jose~R Vazquez-Canteli},
  \bibinfo{person}{Sourav Dey}, \bibinfo{person}{Gregor Henze}, {and}
  \bibinfo{person}{Zoltan Nagy}.} \bibinfo{year}{2020}\natexlab{a}.
\newblock \showarticletitle{{CityLearn: Standardizing Research in Multi-Agent
  Reinforcement Learning for Demand Response and Urban Energy Management}}.
\newblock \bibinfo{journal}{\emph{arXiv}} (\bibinfo{date}{12}
  \bibinfo{year}{2020}).
\newblock
\showISSN{23318422}
\urldef\tempurl%
\url{http://arxiv.org/abs/2012.10504}
\showURL{%
\tempurl}


\bibitem[\protect\citeauthoryear{Vazquez-Canteli, Henze, and
  Nagy}{Vazquez-Canteli et~al\mbox{.}}{2020b}]%
        {Vazquez-Canteli2020}
\bibfield{author}{\bibinfo{person}{Jose~R. Vazquez-Canteli},
  \bibinfo{person}{Gregor Henze}, {and} \bibinfo{person}{Zoltan Nagy}.}
  \bibinfo{year}{2020}\natexlab{b}.
\newblock \showarticletitle{{MARLISA: Multi-Agent Reinforcement Learning with
  Iterative Sequential Action Selection for Load Shaping of Grid-Interactive
  Connected Buildings}}.
\newblock \bibinfo{journal}{\emph{BuildSys 2020 - Proceedings of the 7th ACM
  International Conference on Systems for Energy-Efficient Buildings, Cities,
  and Transportation}} (\bibinfo{year}{2020}), \bibinfo{pages}{170--179}.
\newblock
\showISBNx{9781450380614}
\urldef\tempurl%
\url{https://doi.org/10.1145/3408308.3427604}
\showDOI{\tempurl}


\bibitem[\protect\citeauthoryear{Vazquez-Canteli and Nagy}{Vazquez-Canteli and
  Nagy}{2019}]%
        {Vazquez-Canteli2018d}
\bibfield{author}{\bibinfo{person}{Jose~R. Vazquez-Canteli} {and}
  \bibinfo{person}{Zoltan Nagy}.} \bibinfo{year}{2019}\natexlab{}.
\newblock \showarticletitle{{Reinforcement learning for demand response: A
  review of algorithms and modeling techniques}}.
\newblock \bibinfo{journal}{\emph{Applied Energy}}  \bibinfo{volume}{235}
  (\bibinfo{date}{2} \bibinfo{year}{2019}), \bibinfo{pages}{1072--1089}.
\newblock
\urldef\tempurl%
\url{https://doi.org/10.1016/j.apenergy.2018.11.002}
\showDOI{\tempurl}


\bibitem[\protect\citeauthoryear{Wang and Hong}{Wang and Hong}{2020}]%
        {Wang2020}
\bibfield{author}{\bibinfo{person}{Zhe Wang} {and} \bibinfo{person}{Tianzhen
  Hong}.} \bibinfo{year}{2020}\natexlab{}.
\newblock \showarticletitle{{Reinforcement learning for building controls: The
  opportunities and challenges}}.
\newblock \bibinfo{journal}{\emph{Applied Energy}} \bibinfo{volume}{269},
  \bibinfo{number}{April} (\bibinfo{year}{2020}), \bibinfo{pages}{115036}.
\newblock
\showISSN{03062619}
\urldef\tempurl%
\url{https://doi.org/10.1016/j.apenergy.2020.115036}
\showDOI{\tempurl}


\bibitem[\protect\citeauthoryear{Watkins and Dayan}{Watkins and Dayan}{1992}]%
        {Watkins1992}
\bibfield{author}{\bibinfo{person}{Christopher Watkins} {and}
  \bibinfo{person}{Peter Dayan}.} \bibinfo{year}{1992}\natexlab{}.
\newblock \showarticletitle{{Technical Note: Q-Learning}}.
\newblock \bibinfo{journal}{\emph{Machine Learning}} \bibinfo{volume}{8},
  \bibinfo{number}{3} (\bibinfo{year}{1992}), \bibinfo{pages}{279--292}.
\newblock
\showISBNx{978-1-4613-6608-9}
\showISSN{15730565}
\urldef\tempurl%
\url{https://doi.org/10.1023/A:1022676722315}
\showDOI{\tempurl}


\bibitem[\protect\citeauthoryear{W{\"{o}}lfle, Vishwanath, and
  Schmeck}{W{\"{o}}lfle et~al\mbox{.}}{2020}]%
        {Wolfle2020}
\bibfield{author}{\bibinfo{person}{David W{\"{o}}lfle}, \bibinfo{person}{Arun
  Vishwanath}, {and} \bibinfo{person}{Hartmut Schmeck}.}
  \bibinfo{year}{2020}\natexlab{}.
\newblock \showarticletitle{{A Guide for the Design of Benchmark Environments
  for Building Energy Optimization}}.
\newblock \bibinfo{journal}{\emph{BuildSys 2020 - Proceedings of the 7th ACM
  International Conference on Systems for Energy-Efficient Buildings, Cities,
  and Transportation}} (\bibinfo{year}{2020}), \bibinfo{pages}{220--229}.
\newblock
\showISBNx{9781450380614}
\urldef\tempurl%
\url{https://doi.org/10.1145/3408308.3427614}
\showDOI{\tempurl}


\bibitem[\protect\citeauthoryear{Zhang and Ardakanian}{Zhang and
  Ardakanian}{2020}]%
        {Zhang2020a}
\bibfield{author}{\bibinfo{person}{Tianyu Zhang} {and} \bibinfo{person}{Omid
  Ardakanian}.} \bibinfo{year}{2020}\natexlab{}.
\newblock \showarticletitle{{COBS: COmprehensive Building Simulator}}. In
  \bibinfo{booktitle}{\emph{Proceedings of the 7th ACM International Conference
  on Systems for Energy-Efficient Buildings, Cities, and Transportation}}.
  \bibinfo{publisher}{ACM}, \bibinfo{address}{New York, NY, USA},
  \bibinfo{pages}{314--315}.
\newblock
\showISBNx{9781450380614}
\urldef\tempurl%
\url{https://doi.org/10.1145/3408308.3431119}
\showDOI{\tempurl}


\end{thebibliography}

\end{document}